\documentclass[letterpaper, 10 pt, conference]{ieeeconf}  % Comment this line out if you need a4paper

\IEEEoverridecommandlockouts                              % This command is only needed if 
\usepackage{graphics} % for pdf, bitmapped graphics files
\usepackage{epsfig} % for postscript graphics files
\usepackage{mathptmx} % assumes new font selection scheme installed
\usepackage{times} % assumes new font selection scheme installed
\usepackage{amsmath} % assumes amsmath package installed
\usepackage{amssymb}  % assumes amsmath package installed

\title{\LARGE \bf
GEFM: Graph-Enhanced EEG Foundation Model
}

\author{Limin Wang$^{1}$, Toyotaro Suzumura$^{2}$ and Hiroki Kanezashi$^{3}$% <-this % stops a space
\thanks{$^{1}$Limin Wang ({\tt\small wang-limin516@g.ecc.u-tokyo.ac.jp}), $^{2}$Toyotaro Suzumura ({\tt\small suzumura@acm.org}) and $^{3}$Hiroki Kanezashi ({\tt\small hkanezashi@acm.org}) are with % Graduate School of Information Science and Technology
        The University of Tokyo, 7-3-1 Hongo, Tokyo, Japan
        }%
}
\begin{document}

\maketitle
\thispagestyle{empty}
\pagestyle{empty}

%%%%%%%%%%%%%%%%%%%%%%%%%%%%%%%%%%%%%%%%%%%%%%%%%%%%%%%%%%%%%%%%%%%%
\begin{abstract}

Electroencephalography (EEG) signals provide critical insights for applications in disease diagnosis and healthcare.
However, the scarcity of labeled EEG data poses a significant challenge.
Foundation models offer a promising solution by leveraging large-scale unlabeled data through pre-training, enabling strong performance across diverse tasks.
While both temporal dynamics and inter-channel relationships are vital for understanding EEG signals, existing EEG foundation models primarily focus on the former, overlooking the latter.
To address this limitation, we propose Graph-Enhanced EEG Foundation Model (GEFM), a novel foundation model for EEG that integrates both temporal and inter-channel information. 
Our architecture combines Graph Neural Networks (GNNs), which effectively capture relational structures, with a masked autoencoder to enable efficient pre-training.
We evaluated our approach using three downstream tasks and experimented with various GNN architectures. The results demonstrate that our proposed model, particularly when employing the GCN architecture with optimized configurations, consistently outperformed baseline methods across all tasks.
These findings suggest that our model serves as a robust foundation model for EEG analysis.
\end{abstract}

%%%%%%%%%%%%%%%%%%%%%%%%%%%%%%%%%%%%%%%%%%%%%%%%%%%%%%%%%%%%%%%%%%%%%%%%%%%%%%%%

\section{Introduction}
% Background
% The Importance of EEG Analysis
Understanding brain signals is crucial for clinical diagnosis, neurological disorder prediction, and exploring human cognition.
The analysis of these signals facilitates early disease detection and prevention, forming the foundation of critical healthcare applications.
% Among various brain activity measurement techniques, 
Electroencephalography (EEG) is a widely used method for brain activity measurement, with its analysis techniques undergoing continuous advancements.
Recent progress in machine learning have significantly enhanced EEG analysis, employing models ranging from traditional approaches like support vector machines to advanced architectures such as Transformers and Graph Neural Networks (GNNs).

% In particular, the ability to analyze and interpret brain signals enables critical healthcare applications, such as the early detection of neurological conditions and the prevention of diseases, ultimately promoting healthier lifestyles and improving clinical outcomes.
% As a result, research on brain signals has garnered significant scientific attention in recent years. Among the various techniques developed for measuring brain activity, electroencephalography (EEG) stands out as a widely used and continually advancing method.
% Advancements in machine learning have enabled the application of diverse models to EEG analysis, from traditional methods like support vector machines (SVM) to modern architectures such as Transformers and Graph Neural Networks (GNNs).

% % Advances in Machine Learning Applied to EEG
% Over the years, advancements in machine learning have facilitated the development of diverse approaches for EEG analysis. 
% From traditional methods like feature extraction and support vector machines (SVM) to modern neural network architectures such as Long Short-Term Memory (LSTM) networks, Transformers, and Graph Neural Networks (GNNs), a wide range of models have been explored to enhance the interpretation of EEG data.

% Emergence of Foundation Models
Nowadays, the success of large language models (LLMs) has driven the development and widespread adoption of foundation models. 
These models leverage extensive amounts of unlabeled data during pre-training, achieving strong performance across a wide range of downstream tasks with minimal reliance on labeled data.
Furthermore, foundation models eliminate the need for training from scratch for each task, significantly reducing computational time and cost.
A notable characteristic of foundation models is their ``emergent ability," where larger architectures trained on vast datasets exhibit enhanced generalization capabilities. 
As a result, foundation models have gained traction across various domains, including language~\cite{touvron2023llama, touvron2023llama2}, image~\cite{wang2023internimage, yuan2021florence}, % video, and time-series data,
and are now being explored for EEG analysis.

Foundation models offer significant advantages for EEG applications. 
The scarcity of labeled EEG data, due to the high cost and expertise required for manual annotation, contrasts with the abundance of unlabeled data.
By leveraging this unlabeled data during pre-training, foundation models can be fine-tuned for specific tasks with minimal reliance on labeled samples. 
% EEG research encompasses a wide range of downstream tasks, and as a rapidly evolving field, new research directions and application scenario are continuously emerging.
% Consequently, there is a strong motivation to develop a single, versatile model that can efficiently support diverse EEG tasks, thereby contributing to the advancement of the field.

% Existing Foundation Models in EEG
Several foundation models for EEG analysis, such as BENDR~\cite{kostas2021bendr} and % Brant~\cite{zhang2024brant}, 
Neuro-GPT~\cite{cui2024neuro}, have been proposed, primarily treating EEG as a collection of time-series data for individual channels (electrodes).
These models often rely on advanced sequence modeling techniques, including Transformers and autoencoders, to process temporal information.
However, EEG signals inherently involve complex interactions between channels, making it essential to incorporate inter-channel relationships into the modeling process. 
For instance, abnormalities in these interactions can serve as indicators of neurological disorders, highlighting the potential diagnostic value of capturing such relationships.
Despite this, existing research has not yet adequately addressed the importance of modeling contextual relationships across EEG channels.

% Hypothesis: Developing a GNN-Integrated Foundation Model for EEG
In this study, we propose Graph-Enhanced EEG Foundation Model (GEFM), a novel approach that integrates GNNs into a Transformer-based EEG foundation model to effectively capture inter-channel relationships.
GNNs are particularly well-suited for modeling and representing complex relationships between entities, making them an ideal choice for capturing the intricate interactions among EEG channels.
While the integration of GNNs with sequence modeling techniques, such as combining knowledge graphs (KGs) with LLMs, has been extensively explored in other fields, to the best of our knowledge, this study is the first to apply this combination to EEG analysis.
This approach aims to establish a versatile and task-agnostic foundation model, offering significant potential for advancing EEG research and its applications.

% Detail architecture
To realize our proposed approach, we build upon BENDR~\cite{kostas2021bendr}, a masked autoencoder-based model recognized for its strong performance and adaptability for customization.
In our architecture, we integrate GNNs to enhance the model's ability to capture inter-channel relationships by representing EEG data as a graph structure.
Specifically, each EEG channel is treated as a node, the signals from the corresponding channels serve as node features, and the connections between channels are modeled as edges.

While this integration offers significant potential, it also introduces a key technical challenge.
As a foundation model, the architecture must be applicable across diverse datasets.
However, GNNs typically require fixed-length node features, whereas EEG datasets often vary in signal sequence lengths due to varying task objectives or experimental setups.
This inconsistency makes it impractical to directly use raw EEG signals as node features across all datasets.
To address this, we introduce a sequence length adjustment mechanism that standardizes EEG signal lengths to a predefined target length before they are fed into the GNNs.
% Datasets that already match the target length remain unchanged, ensuring minimal alterations to the data. 
This ensures compatibility across datasets while maintaining model's flexibility as a versatile foundation model for EEG analysis.

% Research Questions
This study aims to address the following research questions:
\begin{description}
    \item[RQ1.] Which GNN architectures demonstrate the best performance in the proposed framework?
    \item[RQ2.] For which types of downstream tasks does the integration of GNNs and sequence modeling techniques provide the most significant improvements?
    \item[RQ3.] What is the most effective strategy for adjusting sequence lengths of EEG signals within the proposed framework?
    \item[RQ4.] Which base model, when combined with GNNs, yields the most robust and versatile foundation model for EEG analysis?
\end{description}
    % \item[RQ4.] What is the most effective way to incorporate inter-channel graphs and GNNs into the foundation models?
    % \item[RQ5.] Why does this approach benefit?
% \end{description}

% Experiments
To address the research questions, we conducted experiments using three GNN architectures: Graph Convolutional Networks (GCN)~\cite{kipf2016semi}, Graph Attention Networks (GAT)~\cite{velivckovic2017graph}, and GraphSAGE~\cite{hamilton2017inductive}. 
The models were pre-trained on a large-scale dataset and evaluated on three downstream tasks to assess their performance. 
To account for the variability in EEG sequence lengths, we examined two adjustment strategies.
% : (1) inserting a linear layer and (2) applying padding to adjust the sequence lengths. 
Furthermore, we tested of two configurations of BENDR~\cite{kostas2021bendr}, which have demonstrated high accuracy in prior studies, as base models integrated with GNNs to develop a robust foundation model for EEG analysis.
The results show that our proposed model, particularly when employing the GCN architecture with optimized configurations, consistently outperformed baseline methods across all tasks.

% Contributions
The contributions of this paper are as follows:
\begin{itemize}
    \item We propose a novel foundation model specifically designed for EEG analysis.
    \item We integrate a robust sequence modeling techniques with GNNs to effectively capture both temporal dynamics and inter-channel relationships in EEG data.
    \item We conduct extensive experimental evaluations across three downstream tasks using three GNN architectures and multiple configurations to identify the optimal setup, demonstrating the effectiveness and versatility of our model.
\end{itemize}

% Our future work ?

\section{Background and Related Work}

\subsection{Foundation Models}
Foundation models are large-scale pre-trained models designed to serve as general-purpose frameworks across a wide range of downstream tasks. 
These models are typically trained on extensive datasets using self-supervised learning, enabling the use of large amounts of unlabeled data.
By leveraging their learned representations, foundation models can be adapted to specific tasks with minimal fine-tuning, demonstrating strong performance and versatility.
This approach reduces the time and cost associated with task-specific model development.
Foundation models have shown significant potential across various domains, including natural language processing (NLP)~\cite{touvron2023llama, touvron2023llama2} and computer vision (CV)~\cite{wang2023internimage, yuan2021florence}.

\subsection{EEG Foundation Models}
Foundation models have recently been introduced in the EEG field to address key challenges, such as the difficulty of large-scale data labeling.
Among these, BENDR~\cite{kostas2021bendr} is a notable example, leveraging techniques inspired by BERT~\cite{devlin2018bert} and wav2vec~\cite{baevski2020wav2vec}.
First, raw multi-channel EEG signals are passed through six convolutional layers (referred to as the ``BENDR Encoder") to generate convolved features.
A portion of the convolved features is then masked, and a Transformer encoder reconstructs the masked features using the unmasked features as context.
The model is optimized by calculating the contrastive loss between the reconstructed features and the original convolved features before masking.

MAEEG~\cite{chien2022maeeg} is an improved model that builds upon the BENDR~\cite{kostas2021bendr} method, achieving enhanced performance. Specifically, a linear layer and a convolutional layer are added after the Transformer encoder to directly reconstruct the EEG signal.
And then instead of computing the contrastive loss between the original features and the reconstructed features, MAEEG calculates the reconstruction loss between the input EEG signal and the reconstructed EEG signal.

Neuro-GPT~\cite{cui2024neuro} is another model that has achieved higher performance than BENDR~\cite{kostas2021bendr}. This model is based on the Generative Pre-trained Transformer (GPT)~\cite{radford2019language} method and uses a decoder-only Transformer Encoder for the encoded features by a new EEG Encoder, which consists of a convolutional module and a Transformer Encoder.

However, to the best of our knowledge, existing EEG foundation models focus exclusively on learning time-series information, while neglecting inter-channel relationships, which are also critical for capturing the underlying dynamics of EEG signals.

\subsection{EEG Models with Graphs}
Several EEG models have been developed to capture inter-channel relationships, although these are not foundation models but are instead trained from scratch for specific tasks.
These models represent EEG signals as graphs, allowing the study of network properties such as channel connectivity.
To process these graph-structured data, they employ Graph Neural Networks (GNNs), which are specifically designed for such applications~\cite{klepl2024graph}.
One notable example is EEG-GCNN~\cite{wagh2020eeg-gcnn}, which utilizes a Graph Convolutional Network (GCN) ~\cite{kipf2016semi} architecture with edge weights calculated based on spatial distances.
This model learns inter-channel relationships effectively by leveraging these edge weights.

\subsection{Graph-Enhanced Models in Other Domains}
While no foundation model in the EEG domain has been developed to learn both time-series information and inter-channel relationships, similar approaches have been explored in the domain of NLP.

% Pre-trained language models (LMs) like BERT~\cite{devlin2018bert} have demonstrated exceptional performance across various NLP tasks.
% These models are typically trained using self-supervised learning techniques, such as masked language modeling, during pre-training.
% Building upon such models, LinkBERT~\cite{yasunaga2022linkbert}, which targets at pre-training LMs with a large scale corpus of documents, regards the corpus as a document graph.
% This study extends the input context beyond a single document by modeling links between documents.

% This approach aligns with our interest in treating data as a graph structure, where node features are processed with powerful architectures while also learning the connectivity between nodes.
% LinkBERT achieves this by introducing an additional pretext task during pre-training: predicting whether two document segments are linked (in the same context) or not.

% However, this method is not directly applicable to EEG data, as EEG signals exhibit less variation compared to textual data, making it difficult to design a similar pretext task for identifying relationships between signals.
% Instead, we propose the use of GNNs, which can effectively model connectivity based on spatial information and are better suited to learning inter-channel relationships in EEG.

% Another 
One example is the pre-training language models (LM) with knowledge graphs (KGs), such as K-BERT~\cite{liu2020k-bert}, which incorporates knowledge graphs into the pre-training process to enhance the model's understanding of entities and relationships.
Some research has proposed the combination of LMs and GNNs to jointly learn on texts and KGs \cite{sun2020colake, yasunaga2021qa,zhang2022greaselm}.
We can apply this approach of integrating GNNs with LMs to EEG data, where the GNNs can capture the relationships between EEG channels, while the LMs can be replaced with Transformer-based sequence models to learn the temporal dynamics of the signals.

% \section{Preliminary}

\section{Methods}
We propose a foundation model that learns both the inter-channel relationships and the time-series dynamics of EEG signals.
Since existing EEG foundation models primarily focus on learning time-series information, we extend these models by integrating inter-channel relationship learning.
Among the state-of-the-art EEG foundation models, BENDR~\cite{kostas2021bendr} is notable for its strong performance and adaptability for customization.
Therefore, we adopt BENDR as the base foundation model for our proposed approach.
In this section, we first provide a brief overview of BENDR, followed by the description of our proposed method.

\begin{figure*}[t]
    \centering
    \includegraphics[width=\linewidth]{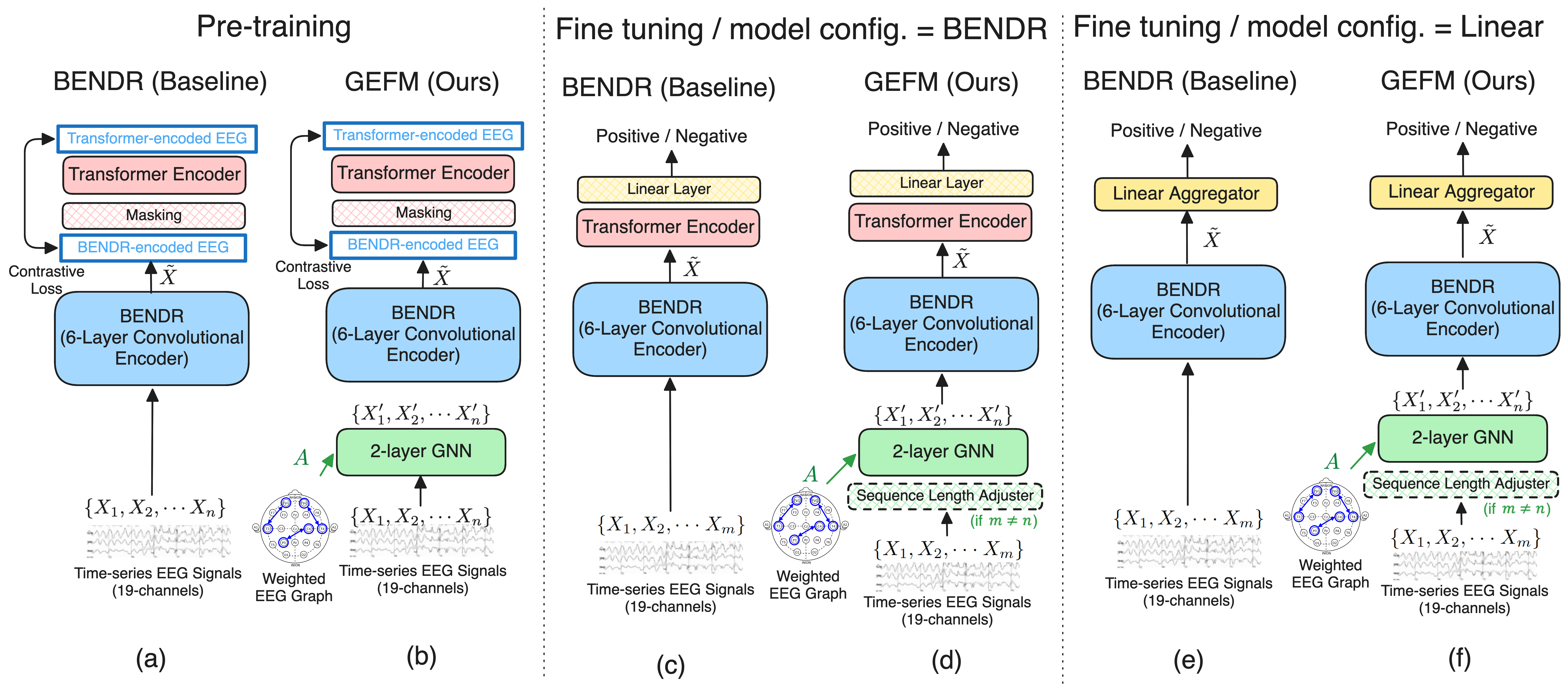}
    \caption{Comparison between two model architectures during pre-training and fine tuning. 
    Figures (a), (c) and (e) represent BENDR~\cite{kostas2021bendr}, while (b), (d) and (f) represent our proposed model, GEFM.
    }
    \label{fig:architectures}
\end{figure*}

\subsection{BENDR}
The model architecture of BENDR is illustrated in (a), (c) and (e) in Figure~\ref{fig:architectures}.
% The model architecture of BENDR is depicted in all the left figures in Figures~\ref{fig:pre-training}, \ref{fig:fine-tuning-BENDR}, \ref{fig:fine-tuning-Linear}.

During pre-training, raw multi-channel EEG signals are passed through six convolutional layers (referred to as the ``BENDR Encoder") to generate convolved features.
The procedure is inspired by wav2vec~\cite{baevski2020wav2vec}.
% Each convolutional layer conducts Dropout and GroupNorm, and the output activation function is GELU.
A portion of the convolved features is then masked, and a Transformer encoder reconstructs the masked features using the unmasked features as context.
The model is optimized by calculating the contrastive loss between the reconstructed features and the original convolved features before masking.

For downstream tasks, several model configurations have been proposed.
Among these, we focus on two configurations (``BENDR" and ``Linear") that achieved the highest performance according to \cite{kostas2021bendr}.
In both configurations, the process up to encoding by the ``BENDR Encoder" is identical to that of the pre-training phase.
However, the subsequent steps differ.
In the ``BENDR" configuration, the convolved features are passed to the pre-trained Transformer encoder, followed by a linear layer % with softmax activation 
for classification.
In contrast, in the ``Linear" configuration, the pre-trained Transformer encoder is omitted entirely, and the convolved features are aggregated and passed to a linear classification layer.

\subsection{Our Proposal: GEFM}
We propose Graph-Enhanced EEG Foundation Model (GEFM), extending the BENDR architecture described above by incorporating the learning of inter-channel relationships.
Inter-channel relationships in EEG signals can be naturally represented using a graph structure, where each channel corresponds to a node and the edges represent the connectivity between channels.
Graph Neural Networks (GNNs) are particularly well-suited for modeling such relationships in graph-structured data.
Therefore, we propose integrating GNNs with BENDR to enhance its capability of learning inter-channel relationships.
% Therefore, we propose integrating GNNs with foundation models that primarily focus on learning time-series information, particularly in combination with BENDR.

To achieve this, we employ a simple yet effective strategy by inserting a two-layer GNN directly before the inputs are processed by the BENDR Encoder, in both the pre-training and downstream tasks.
The GNN assumes an input graph structure $G = (V, E)$ defined as follows:
\begin{itemize}
    \item Each node in $V$ represents an EEG recording channel (electrode), where $|V| = C$, corresponding to the total number of channels.
    % That is, when the EEG recordings are in $C$ channels, $V$ is the set of $|V| = M$ channels.
    \item The node features are the EEG recordings for each channel, with a feature length equal to the sequence length $n$ of the recordings.
    \item Following EEG-GCNN~\cite{wagh2020eeg-gcnn}, the graph is fully connected, and each edge $(u, v) \in E$ has a weight. This setup enables the GNN to capture all pairwise connectivity between channels. The edge weights are represented as a matrix $W \in \mathbf{R} ^ {C \times C}$.
    \item The edge weights are defined as the reciprocal of the geodesic distance between two channels on the spherical scalp model, based on the hypothesis that closer channels interact more strongly. The geodesic distance $D_{ij}$ between channels $i$ and $j$ is computed using their 3D coordinates $(x_i, y_i, z_i)$ and $(x_j, y_j, z_j)$, along with the sphere's radius $r$, as:
    \begin{eqnarray}
        D_{ij} = \arccos\left(\frac{x_i x_j + y_i y_j + z_i z_j}{r^2}\right) \nonumber
    \end{eqnarray}
    The coordinates and radius values are adopted from EEG-GCNN. These edge weights are utilized by the GNN to learn inter-channel relationships effectively.
\end{itemize}

% The GNN models can be either of GCN, GAT or GraphSAGE.
% In the case of GCN, the equation of the message passing will be \ref{eq:GCN}

% \begin{equation}\label{eq:GCN}
%     h_i^{(l+1)} = \sigma (W_1^{(l)}~h_i^{(l)} + \sum_{j \in N(v_i)}{}{W_2{(l)}~h_j{(l)}~e_{ji}})
% \end{equation}

% \begin{equation}
%     \{X'_1, X'_2,\cdot\cdot\cdot~X'_N \} \nonumber
% \end{equation}

% \begin{equation}
%     \{X_1, X_2,\cdot\cdot\cdot~X_N \} \nonumber
% \end{equation}

% \begin{equation}
%     \{X_1, X_2, \cdot\cdot\cdot~X_m \} \nonumber 
% \end{equation}

% $
% While this integration offers significant potential, it also introduces a key technical challenge.
% GNNs typically require fixed-length node features, yet EEG datasets often exhibit variability in signal sequence lengths due to varying task objectives or experimental setups.
% This inconsistency makes it infeasible to directly use raw EEG signals as node features across all datasets.
% % 
% To address this issue, we align the sequence lengths of EEG signals across datasets by adjusting them to a predefined target length before feeding them into the GNN.
% % Datasets that already match the target length remain unchanged, ensuring minimal alterations to the data. 
% This ensures compatibility across diverse datasets while maintaining model's flexibility as a versatile foundation model for EEG analysis.
% $

Our proposed architecture is illustrated in (b), (d) and (f) in Figure~\ref{fig:architectures}.
% are shown in all the right figures in Figures~\ref{fig:pre-training}, \ref{fig:fine-tuning-BENDR}, \ref{fig:fine-tuning-Linear}.
The GNNs in this architecture are pre-trained and subsequently fine-tuned for downstream tasks.
The process following the ``BENDR Encoder" remains identical to that of the base model, including the use of two model configurations, ``BENDR" and ``Linear", for downstream tasks.
% - GNNはpre-trainした後再利用しfine tuningする。
% - ``BENDR Encoder"以降の過程はbase modelの時と同様である。またbase modelと同様、downstream taskにおいて、``BENDR", ``Linear"の2つのmodel configurationを使用する。
% - foundation modelとして、このアーキテクチャ/パラメータは、共通のものをさまざまなデータセットで使うことを想定している。

As a foundation model, this architecture must be applicable across diverse datasets.
However, a key challenge arises because GNNs require fixed-length node features, whereas EEG datasets often vary in signal sequence lengths due to differing task objectives. 
To address this, we introduce a sequence length adjustment mechanism, such as padding or utilizing a linear layer, that standardizes EEG signal sequence lengths to a predefined target. 
This ensures compatibility across datasets while maintaining the model's versatility for EEG analysis.
If the original sequence length is already equal to the predefined target, no adjustment is applied to avoid unnecessary data manipulation.

\subsection{Alternative Architectural Variants}
Our proposed method aims to learn both temporal and spatial information in EEG signals as a foundation model. To achieve this, we leverage the strengths of existing EEG foundation models in learning temporal information and add a module to learn spatial relationships between channels.

There are three possible ways to combine these modules: 
\begin{enumerate}
    \item\label{item:variant1} Learning spatial relationships first and then temporal information (This is the method we adopted, as described above).
    \item\label{item:variant2} Learning temporal information first and then spatial relationships.
    \item\label{item:variant3} Learning both in parallel and aggregating the results.
\end{enumerate}

However, approach~\ref{item:variant2} is not suitable for our purpose because it would require placing the GNN after the BENDR Encoder and Transformer Encoder, which would integrate all channel information and make it difficult to learn spatial relationships.

Approach~\ref{item:variant3} is thought to be feasible, but we chose to focus on approach~\ref{item:variant1} in this study because it allows us to reuse the BENDR Encoder and pretext task with minimal modifications. Approach~\ref{item:variant3} is worth exploring in future research.

In this study, we adopt approach~\ref{item:variant1}, where we first learn spatial relationships using a GNN and then learn temporal information using the BENDR Encoder and Transformer Encoder. This approach enables us to effectively learn both spatial and temporal information in EEG signals.

\subsection{Incorporating Graphs Into Other EEG Foundation Models}
In addition to BENDR~\cite{kostas2021bendr}, there also exist other EEG foundation models.
Although we did not conduct experiments on them, we believe that a similar approach can be applied to them as well. 
Here we provide a brief overview of how our proposed method can be applied to some other EEG foundation models, such as MAEEG~\cite{chien2022maeeg} and Neuro-GPT~\cite{cui2024neuro}.

% For MAEEG~\cite{chien2022maeeg}, we can construct a graph by treating each channel of the multi-channel EEG data as a node. As shown in Figure~\ref{fig:MAEEG}, the 6-layer convolutional encoder integrates information from all channels. By inserting a GNN before this encoder, we can learn the relationships between channels and potentially improve the performance of MAEEG.

% Similarly, for Neuro-GPT~\cite{cui2024neuro}, we can construct a graph by treating each channel as a node. As shown in Figure~\ref{fig:Neuro-GPT}, the EEG Encoder first uses a Convolutional Module to convolve information from all channels, and then uses a Transformer Module to incorporate temporal information. We propose inserting a GNN before the Convolutional Module to learn the relationships between channels.

Both models first use a Convolutional Encoder to process the input EEG signals and then use a Transformer Encoder to capture temporal information.
We consider a simple yet effective method, inserting a GNN before the Convolutional Encoder to learn the relationships between channels, will be effective for these models.
By incorporating graph structures into these EEG foundation models, we can potentially improve their performance and better capture the complex relationships between channels in EEG data.

\section{Experiments}
We conducted experiments, including both pre-training and downstream tasks, under various conditions to evaluate the performance of GEFM and gain insights into the research questions.

\subsection{Dataset}

\subsubsection{Pre-training Dataset}
Following BENDR~\cite{kostas2021bendr}, we utilized the Temple University Hospital EEG Corpus (TUEG)~\cite{obeid2016temple} for pre-training.
TUEG provides a diverse range of subjects and includes recordings across multiple sessions over extended time periods, making it an ideal dataset for pre-training foundation models.
For our study, we specifically focused on version 2 of this dataset, which consists of clinical recordings from over 10,000 individuals.
% and amounts to approximately ?TB of European-data-format (EDF) EEG recordings.
To accelerate experimentation and evaluation during development, we downsampled the dataset to one-tenth of its original size.

% preprocessing

\subsubsection{Downstream Datasets}
We evaluated GEFM using the datasets from the following tasks.

\begin{description}
    \item[MMI]\cite{goldberger2000physiobank, schalk2004bci2000} This task involves predicting whether the participant is imagining the movement of the right (positive) or left (negative) hand.
    \item[P300]\cite{goldberger2000physiobank, citi2010documenting} This task involves predicting whether the participant focused on a flashing target letter (positive) or a non-target letter (negative).
    \item[ERN]\cite{margaux2012objective} This task involves predicting whether the participant's attempt to input a character using a P300 speller was recognized correctly (positive) or incorrectly (negative).
\end{description}

All three tasks are binary classification problems, which were previously used in BENDR~\cite{kostas2021bendr}.
These tasks involve EEG signals recorded from a sufficient number of channels in the 10/20 channel scheme~\cite{jurcak200710}, making them suitable for constructing graphs for GNN-based approaches.
Detailed information about these datasets is provided in Table~\ref{tab:downstream_datasets}.
% related to brain-computer interface (BCI) applications.

The P300 and ERN datasets exhibit imbalanced class distributions, while the MMI dataset is balanced.
To address class imbalance during fine tuning, we followed the methodology presented in BENDR~\cite{kostas2021bendr} and applied the following steps:
\begin{itemize}
    \item For imbalanced datasets, during training we performed undersampling of the majority class to equalize the number of samples across classes.
    \item During testing, we evaluated performance using metrics that account for class imbalance, specifically AUROC for P300 and ERN. For MMI, which has balanced classes, we used Accuracy to assess test performance.
\end{itemize}

% %% EGA の有効性を比較するために、他の論文で使われている downstream tasks で比較検証し、その上でヘルスケアの downstream task で検証するために TUAB タスクで評価した
% To evaluate the performance of EEG-GraphAdapter, we conducted model performance evaluations across four binary classification downstream tasks as shown in Table \ref{tb:ds-tasks} with various domains, EEG sequence lengths, and label distributions. The MMI\cite{mmi1,mmi2}, P300 \cite{p300}, and ERN \cite{ern} tasks are identical to those used in the baseline BENDR model. These evaluations were designed to address the corresponding research questions for each task.

% To evaluate the performance of EGA, we conducted model performance evaluations on two binary classification tasks (MDD, TUAB) related to healthcare as downstream tasks, as shown in Table \ref{tab:downstream_datasets}.
% We used datasets \cite{mdd-data,tuab-data} that are publicly available and have explicit labels for abnormal (positive) and healthy (negative) samples for downstream tasks.
% % To evaluate the performance of EGA, we conducted model performance evaluations on three binary classification downstream tasks, as shown in Table \ref{tb:ds-tasks}. These tasks include MMI\cite{mmi1} and P300\cite{p300}, which are commonly used in BCI-related evaluations, such as in the baseline BENDR model. We also evaluated the performance in the TUAB\cite{tuab-data} downstream task to evaluate the model's applicability to healthcare-related downstream tasks.

\begin{table*}[t]
    \centering
    \begin{tabular}{lccccc}
    \hline\hline
    \textbf{Dataset} & \textbf{sfreq. (Hz)} & \textbf{Length (s)} & \textbf{Num of Ch.} & \textbf{Subjects} & \textbf{Folds} \\
    \hline
    \textbf{MMI}~\cite{goldberger2000physiobank, schalk2004bci2000} & 160 & 6 & 64 & 105 & 5 \\
    \textbf{P300}~\cite{goldberger2000physiobank, citi2010documenting} & 2,048 & 2 & 64 & 9 & 9 \\
    \textbf{ERN}~\cite{margaux2012objective} & 200 & 2 & 56 & 26(10) & 4 \\
    % TUAB & 
    \hline\hline
    \end{tabular}
    \caption{Downstream dataset battery and number of cross-validation folds used, following BENDR~\cite{kostas2021bendr}.}
    \label{tab:downstream_datasets}
\end{table*}

\subsubsection{Preprocessing}
We applied the following preprocessing steps to both the pre-training and downstream datasets, following the methodology presented in BENDR~\cite{kostas2021bendr}:

\begin{itemize}
    \item To standardize the sampling frequency across datasets, we ensured that all recordings had a frequency of 256 Hz by applying over- or undersampling as necessary.
    \item We utilized 19 EEG channels from the 10/20 channel scheme~\cite{jurcak200710} and ignored all other channels.
    \item For pre-training, 60-second sequences were extracted from the pre-training dataset for use in general training, while 20-second sequences were specifically used for the P300-related training, as described in BENDR~\cite{kostas2021bendr}. For downstream tasks, the entire length of the sequences was used.
    % This means that every subsequence holds 15,360 samples., which means that every sequence from the dataset of 6 seconds amounts to 1536 samples, and every sequence from the dataset of 2 seconds amounts to 512 samples.
\end{itemize}

\subsection{Setup}

\subsubsection{GNN architectures}
We evaluated the performance of GEFM by individually incorporating three standard GNN architectures: Graph Convolutional Networks (GCN)~\cite{kipf2016semi}, Graph Attention Netowrks (GAT)~\cite{velivckovic2017graph} and GraphSAGE~\cite{hamilton2017inductive}. 
Each architecture was used consistently throughout a single set of pre-training and downstream tasks.
All GNN implementations were based on PyTorch 2.3.1 and PyTorch Geometric 2.5.3.

Our experiments included five configurations: GCN and GAT with and without edge weights, and GraphSAGE without edge weights.
While GCN and GAT were configured to utilize edge weights, 
using the \texttt{edge\_weight} and \texttt{edge\_attr} inputs, respectively, 
GraphSAGE does not natively support edge weights in its standard implementations in PyTorch Geometric 2.5.3. 
Consequently, GraphSAGE was tested without edge weights.
% GraphSAGE could not process edge weights due to limitations in its standard implementations in PyTorch Geometric 2.5.3. 
To ensure a fair comparison, we also tested GCN and GAT without edge weights. 
% This allowed us to evaluate the relative performance of these architectures under comparable conditions.

% Experiments were conducted with GCN and GAT incorporating edge weights, and without edge weights for comparison.were able to process edge weights described in the previous sections, using the form of \texttt{edge\_weight} input for GCN and the form of \texttt{edge\_attr} input for GAT during prediction.
% However, GraphSAGE wasn't implemented to process edge weights using either the \texttt{edge\_weight} or \texttt{edge\_attr} format.
% Therefore, GraphSAGE ignores edge weights.
% For comparison, we also experimented GCN and GAT without edge weights.

% \subsubsection{RQ2. For what kind of tasks does our proposal show the significant impact?}
% We adapted 3 downstream tasks, which are described in the next subsection, to measure the performance of our proposed model.
% We aim to discuss whether there is difference of performance between tasks and what is the relationship between the difference and the characteristics of tasks.

\subsubsection{Sequence Length Adjusters}
As described in the previous section, all sequence lengths must be standardized to a specific value, $n$, to meet the requirements of a foundation model.
To fully leverage the information available during pre-training, we fixed $n$ to the sequence length of the pre-training dataset.
Consequently, for downstream datasets with sequence lengths $m$, we adjusted $m$ to match $n$. 

Since $m$ is typically smaller than $n$, we explored two simple yet effective adjustment methods.
The first inserts a linear layer of size $m \times n$ immediately before the GNN.
% This linear layer that takes an input of size $m$ and produces an output of size $n$.
The second uses padding, where the last value of the original signal is repeated and appended to the sequence until the sequence reaches $n$.

\subsection{Results}

The results are presented in Tables~\ref{tab:results-linear} and \ref{tab:results-padding}.
% The main results are shown in Figures~\ref{fig:MMI_r}, \ref{fig:P300_r}, \ref{fig:ERN_r}.
The baseline corresponds to the original BENDR~\cite{kostas2021bendr}.
Note that the evaluation metric for MMI is Accuracy, while AUROC is used for P300 and ERN. 
% The abbreviation ``(w/ e.)" in the tables indicates that the GNN utilized edge weights.
Those with the statement ``(with edge weights)" in the tables indicates that the GNN utilized edge weights.
For the experiments with \textit{padding} as the sequence length adjuster (shown in Table~\ref{tab:results-padding}), GCN and GAT without edge weights were excluded from these experiments due to their poor performance with the linear layer (see Table~\ref{tab:results-linear}). 
The following discussion addresses key points related to our research questions.

\subsubsection{Comparison of GNN Architectures (RQ1)}

% GCN, GAT, GraphSAGE
Tables~\ref{tab:results-linear} and \ref{tab:results-padding} indicate that our proposed approach performed better when using a linear layer for the sequence length adjustment compared to padding.
Therefore, to analyze the performance variations introduced by different GNN architectures, the following discussion focuses on the results obtained with a linear layer.

As shown in Table~\ref{tab:results-linear}, among all the GNN architectures tested, only ``GCN with edge weights" consistently outperformed the baseline across all three downstream tasks.
Thus, ``GCN with edge weights" emerges as the most suitable architecture for incorporation into foundation models for EEG analysis.
The next best-performing architecture is ``GraphSAGE", which exceeded the baseline in two out of three tasks, making it another promising candidate.
Investigating the underlying factors contributing to this behavior remains an open question and is expected to be addressed in future work.

\begin{table}
    \centering
    \begin{tabular}{llrrr}
        \hline\hline
        \textbf{Model} & \textbf{Config.} & \textbf{MMI} & \textbf{P300} & \textbf{ERN} \\
        % \textbf{Model Architecture} & \textbf{Model Config.} & \textbf{Accuracy MMI} & \textbf{AUROC on P300} & \textbf{AUROC on ERN} \\
        \hline
        \textbf{Baseline} & BENDR & 0.646 & 0.577 & \textbf{0.522} \\
         & Linear & \textbf{0.794} & \textbf{0.607} & 0.508 \\
        \hline
        \textbf{GraphSAGE} & BENDR & \underline{0.883} & \underline{0.692} & 0.501 \\
         & Linear & 0.758 & 0.580 & 0.492 \\
        \textbf{GCN} & BENDR & 0.514 & \underline{0.616} & \underline{0.534} \\
         & Linear & 0.506 & 0.578 & 0.486 \\
        \textbf{GCN (with edge weights)} & BENDR & \underline{\textbf{0.849}} & \underline{\textbf{0.616}} & \underline{\textbf{0.538}} \\
         & Linear & 0.508 & 0.574 & 0.504 \\
        \textbf{GAT} & BENDR & 0.500 & \underline{0.618} & \underline{0.551} \\
         & Linear & 0.509 & 0.578 & 0.496 \\
        \textbf{GAT (with edge weights)} & BENDR & 0.509 & \underline{0.620} & \underline{0.525} \\
         & Linear & 0.508 & 0.577 & 0.500 \\
        \hline\hline
    \end{tabular}
    \caption{The results for all downstream tasks and GNN architectures using \textit{a linear layer} as the sequence length adjuster. 
    % Entries from our proposed models that exceed the baseline are underlined.
    }
    \label{tab:results-linear}
\end{table}

\subsubsection{Differences of the Effect of GNNs Arising from Downstream Tasks (RQ2)}
% MMI
This section examines how the effect of introducing GNNs varies across downstream tasks by comparing the performance differences between the baseline and GEFM.
Specifically, we evaluate GEFM configured as ``GCN with edge weights", identified as the most suitable architecture in the previous section, with the base model configuration set to ``BENDR" and the sequence length adjuster implemented as \textit{a linear layer}.
For a fair comparison across tasks, we use the same configurations for the baseline.
Hereafter, we refer to `GEFM configured as ``GCN with edge weights"' simply as `GEFM'.

As shown in Table~\ref{tab:results-linear}, on MMI, the baseline achieved a score of $0.646$, while GEFM achieved $0.849$, representing a $31.4\%$ improvement. 
On P300, the baseline achieved $0.568$, with GEFM improving performance by $8.53\%$. 
On ERN, the baseline achieved $0.522$, and GEFM showed a $3.11\%$ improvement. 
These results indicate that the higher the baseline performance on a task, the greater the relative improvement achieved by GEFM.
Further analysis could investigate the relationship between model performance and task-specific characteristics, particularly the physiological features associated with each task, to gain deeper insights.

\subsubsection{Comparison of Sequence Length Adjusters (RQ3)}
% padding vs insert linear layer
Tables~\ref{tab:results-linear}, \ref{tab:results-padding} show that all the models using a linear layer as the sequence length adjuster consistently outperformed those using padding.
Consequently, more models outperformed the baseline when using a linear layer compared to padding.
This indicates that adding a linear layer before GNNs is a more effective method for sequence length adjustment.

One possible explanation for this observation lies in the difference between the sequence lengths before and after adjustment.
The sequence length before adjustment was less than half, and in the smallest cases as small as one-tenth, of that after adjustment. 
When using padding, a significant proportion of the adjusted sequence consisted of newly added padding values, overshadowing the meaningful information from the original signals. 
Consequently, the model struggled to learn effective representations, leading to lower performance.

In contrast, when using a linear layer for sequence length adjustment, the original signal was distributed more sparsely across the adjusted sequence.
While the data was ``stretched", the essential characteristics of the signals were preserved throughout the sequence. 
This allowed the model to capture the critical features of the original data more effectively, enabling the GNNs to leverage these features and achieve better performance.

\begin{table}
    \centering
    \begin{tabular}{llrrr}
        \hline\hline
        \textbf{Model} & \textbf{Config.} & \textbf{MMI} & \textbf{P300} & \textbf{ERN} \\
        \hline
        \textbf{Baseline} & BENDR & 0.646 & 0.568 & \textbf{0.522} \\
         & Linear & \textbf{0.794} & \textbf{0.608} & 0.508 \\ 
        \hline
        \textbf{GraphSAGE} & BENDR & \underline{0.874} & 0.512 & 0.468 \\
         & Linear & 0.538 & 0.503 & 0.475 \\
        \textbf{GCN (with edge weights)} & BENDR & 0.521 & 0.504 & 0.481 \\
         & Linear & 0.505 & 0.508 & 0.471 \\
        \textbf{GAT (with edge weights)} & BENDR & 0.506 & 0.504 & 0.495 \\
         & Linear & 0.502 & 0.493 & 0.490 \\ 
        \hline\hline    
    \end{tabular}
    \caption{The results for all downstream tasks and the GNN architectures we experimented with using \textit{padding} as the sequence length adjuster.
    % GCN and GAT without edge weights were excluded from these experiments due to their poor performance with the linear layer (see Table~\ref{tab:results-linear}). 
    % Entries from our proposed models that exceed the baseline are underlined.
    }
    \label{tab:results-padding}
\end{table}

\subsubsection{Comparison of Base Model Configurations (RQ4)}
This section discusses whether incorporating GNNs is more effective when the base model configuration is ``BENDR" or ``Linear."
As shown in Tables~\ref{tab:results-linear} and \ref{tab:results-padding}, when comparing the performance of models employing GNNs across all task-model combinations, the ``BENDR" configuration outperformed the ``Linear" configuration in 21 out of 24 cases.
In contrast, among the baseline models, ``Linear" exceeded ``BENDR" in 2 out of 3 cases.
Notably, all GNN-based models that outperformed the baseline were configured with ``BENDR."
These results indicate that incorporating GNNs is more effective when the base model configuration is ``BENDR."

A possible explanation for this observation is as follows:
Incorporating GNNs introduces additional information, such as inter-channel relationships, into the feature representation derived from raw signals.
To process and utilize this enriched information effectively, the model requires a greater number of parameters after the GNN layers.
Compared to ``Linear", the ``BENDR" configuration includes more parameters and, more importantly, employs a Transformer Encoder, which is a powerful mechanism for feature extraction.
We hypothesize that these factors enable the model to better leverage the information encoded by the GNNs, resulting in improved performance.

\section{Conclusion}
Foundation models for EEG analysis are particularly valuable due to the difficulty of collecting large amounts of labeled data and their ability to be applied to a wide range of EEG tasks with minimal computational time and cost.
In this study, we propose Graph-Enhanced EEG Foundation Model (GEFM), a novel foundation model for EEG that leverages both inter-channel relationships and the temporal dynamics of EEG signals.
The proposed architecture integrates GNNs, which are effective at learning relationships between entities, with a masked autoencoder-based framework.
We evaluated the model using several GNN architectures across three downstream tasks.
The results indicate that GEFM, when employing the GCN architecture~\cite{kipf2016semi}  with specific configurations, consistently outperformed the baseline across all tasks.
These findings demonstrate that incorporating inter-channel relationships learning through GNNs enhances the model's performance, establishing it as a more effective foundation model for EEG analysis.

% Future Work
As future work, we plan to evaluate GEFM on a broader range of tasks to further demonstrate its versatility. 
Additionally, we aim to investigate the mechanisms underlying the observed improvements achieved through integrating graph structures, using techniques such as GNNExplainer~\cite{ying2019gnnexplainer}. 
And another potential direction is to expand our approach, which integrates inter-channel relationships and time-series information, to other base models or various self-supervised learning approaches.
Alternatively, we may design a novel architecture specifically optimized for EEG foundation models.

\section{Acknowledgments}
This work is partially supported by JSPS KAKENHI Grant JP21K17749 and JP23K28098. 

\bibliographystyle{ieeetr}
\bibliography{reference}

\end{document}